\documentclass[runningheads]{llncs}
\usepackage{hyperref}
\usepackage{subcaption}
\usepackage{tabularx}
\usepackage{listings}
\usepackage{verbatim}
\usepackage[table]{xcolor}
\usepackage{setspace}
\usepackage{enumitem}
\usepackage{float}
\usepackage[]{caption}
\usepackage{orcidlink}
\usepackage{pdfpages}
\usepackage{enumitem}
\usepackage{graphicx}
\usepackage{booktabs}
\usepackage{amssymb}
\usepackage{rotating}
\usepackage{bbm}
\usepackage{array}
\begin{document}

\title{PM-LLM-Benchmark: Evaluating Large Language Models on Process Mining Tasks}
\titlerunning{PM-LLM-Benchmark}

\author{
Alessandro Berti\inst{1,2}\orcidlink{0000-0002-3279-4795},
Humam Kourani\inst{2,1}\orcidlink{0000-0003-2375-2152},
Wil M. P. van der Aalst\inst{1,2}\orcidlink{0000-0002-0955-6940}
}
\authorrunning{A. Berti et al.}
\institute{Process and Data Science Chair, RWTH Aachen University, Aachen, Germany \and
Fraunhofer FIT, Sankt Augustin, Germany \\
\email{\{a.berti, wvdaalst\}@pads.rwth-aachen.de; humam.kourani@fit.fraunhofer.de}}
\maketitle

\begin{abstract}
Large Language Models (LLMs) have the potential to semi-automate some process mining (PM) analyses.
While commercial models are already adequate for many analytics tasks, the competitive level
of open-source LLMs in PM tasks is unknown. In this paper, we propose \emph{PM-LLM-Benchmark}, the first comprehensive benchmark for PM
focusing on domain knowledge (process-mining-specific and process-specific) and on different implementation strategies.
We focus also on the challenges in creating such a benchmark, related to the public availability of the data and on evaluation biases
by the LLMs.
Overall, we observe that most of the considered LLMs can perform some process mining tasks at a satisfactory level,
but tiny models that would run on edge devices are still inadequate.
We also conclude that while the proposed benchmark is useful for identifying LLMs that are adequate for process mining tasks, further research is needed
to overcome the evaluation biases and perform a more thorough ranking of the ``competitive'' LLMs.
\end{abstract}

\renewcommand{\sectionautorefname}{Section}
\renewcommand{\subsectionautorefname}{Section}
\renewcommand{\subsubsectionautorefname}{Section}
\def\univs{U}
\newcommand{\univ}[1]{\univs_{\mathit{#1}}}
\newcommand{\class}[1]{\mathbb{C}_{\mathit{#1}}}
\newcommand{\pim}[1]{\pi_{\mathit{#1}}}

\let\olddefinition\definition
\renewcommand{\definition}{\small\olddefinition}


\section{Introduction}
\label{sec:introduction}

Process mining (PM) is a branch of data science aiming to derive process-related insights from the event data recorded during the execution of a process.
A wide set of automated PM techniques exist for process discovery (the automated discovery of process models starting from the event data),
conformance checking (comparing event data and process models), and model enhancement (annotating a process model with metrics derived from the event data).
PM could benefit significantly from the provision of domain knowledge \cite{DBLP:conf/simpda/DixitBAHB15a}. Modern Large Language Models (LLMs) have the capability to follow
the instructions contained in a given prompt and are trained on large sets of generic knowledge, including process-related knowledge. The release of OpenAI's GPT-4 has been a milestone,
as such LLM proved capable in different PM tasks \cite{DBLP:conf/bpm/Berti0A23} including semantic anomaly detection and root cause analysis. However, GPT-4 is a commercial LLM,
and open-source LLMs were significantly behind the quality of GPT-4. Recently, many companies released good-performing open-source LLMs, which in general-purpose benchmarks approach GPT-4-level
quality\footnote{\url{https://chat.lmsys.org/}}. For example, \emph{Llama 3}\footnote{\url{https://llama.meta.com/llama3/}} by Meta, \emph{Mixtral 8x7B} and \emph{Mixtral 8x22B}\footnote{\url{https://mistral.ai/news/mixtral-of-experts/}} by Mistral, and \emph{WizardLM2}\footnote{\url{https://huggingface.co/WizardLM}} by Microsoft are all-rounder LLMs.
Also alternative commercial models have been proposed from Antrophic (\emph{Claude AI}\footnote{\url{https://claude.ai/}}) and Google (\emph{Gemini\footnote{\url{https://gemini.google.com/}}}),
which also approach GPT-4 levels of quality.

Given the large number of commercial and open-source LLMs, benchmarks are essential to distinguish between good and bad-for-the-purpose LLMs. While many general-purpose benchmarks exist for LLMs, there is a lack of comprehensive benchmarks in process mining (PM). Several factors contribute to the difficulty of proposing such a benchmark: multiple PM artifacts exist (e.g., traditional/object-centric event logs, procedural/declarative process models, situation tables); multiple PM types exist (e.g., process discovery, conformance checking, applications of machine learning such as anomaly detection and root cause analysis, predictive analytics); multiple PM practices exist, with different pathways followed during a process mining analysis \cite{DBLP:conf/bpm/ZerbatoSW21}; multiple PM code libraries and query languages exist, including Python (e.g., pm4py), SQL, and non-relational languages; and the ability to propose valuable answers depends on the ability of the analyst to propose valuable inquiries/hypotheses \cite{DBLP:conf/bpm/Berti0A24}.

In this paper, we propose three main contributions:
i) a first \emph{comprehensive benchmark for process mining tasks executable by LLMs}, focusing on two implementation paradigms (direct provision of insights and code generation), and including several categories of ``static'' prompts (stored in TXT files) requiring process-mining-specific and process-specific domain knowledge;
ii) a \emph{scalable evaluation strategy} to assess the quality of the textual/coding answers provided by LLMs;
iii) the \emph{results of the application of the benchmark} on several state-of-the-art LLMs.
While the benchmark provides a score useful to rank LLMs, some caveats discussed in Section \ref{sec:benchmark} suggest avoiding comparing the scores for highly-performing LLMs. In particular, the role and limitations of LLMs as judges for PM tasks' outputs need to be discussed along with the need for ground truth in scoring the answers. Moreover, advanced implementation paradigms (RAG, agents crew, multi-stage hypothesis generation), discussed in Section \ref{subsec:futureBenchmarkingStrategies}, are not assessed by the benchmark.

The rest of the paper is organized as follows. Section \ref{sec:relatedWork} describes the related work connecting process mining and LLMs, and the evaluation of LLMs outputs.
Section \ref{sec:benchmark} introduces the categories of prompts included in the benchmark and the proposed evaluation strategy.
Section \ref{sec:benchmarkResults} discusses the results of the benchmark on state-of-the-art LLMs.
Section \ref{subsec:futureBenchmarkingStrategies} proposes some novel scenarios requiring novel benchmarks.
Finally, Section \ref{sec:conclusion} concludes the paper.

\section{Related Work}
\label{sec:relatedWork}

\noindent
\textbf{Connecting LLMs to PM}:
Several Business Process Management tasks have been linked with LLMs \cite{DBLP:conf/bpm/VidgofBM23}.
For instance, process modeling exploits LLMs to create process models starting from textual descriptions \cite{DBLP:conf/bpm/GrohsAER23,DBLP:demo/HumamPromoAI,DBLP:conf/bpm/Humam0A24}.
Process mining tasks have been implemented on LLMs thanks to textual description of the mainstream artifacts (event logs, process models) \cite{DBLP:conf/bpm/Berti0A23}.
Three implementation paradigms are identified \cite{DBLP:journals/corr/abs-2307-12701}: i) \emph{direct provision of insights}, ii) \emph{generation of database queries} (SQL), and iii) \emph{autonomous hypotheses generation}.
The direct provision of insights requires the provision of the necessary information to the LLM. Generating database queries \cite{DBLP:journals/corr/abs-2307-09909} uses LLMs to create (SQL) statements to be executed
against the data source, mitigating privacy risks but not exploiting the domain knowledge of LLMs. The autonomous formulation of hypotheses combines the two methodologies allowing the LLM to
create database queries and interpret their output.
In \cite{DBLP:conf/bpm/Berti0A24}, the capabilities required for process mining on LLMs are described: \emph{acceptance of long prompts}, allowing for the provision of a significant amount of information to the LLM; \emph{acceptance of visual prompts}, as visualizations allow us to easily identify process-related patterns; \emph{coding capabilities}, including the generation of scripts and SQL statements; and \emph{factuality}, being able to cross-check the outputs against knowledge bases or search engines. Given the absence of process-mining-specific benchmarks, \cite{DBLP:conf/bpm/Berti0A24} suggests using general-purpose benchmarks (using traditional, domain-knowledge, visual, coding, fairness, and hypotheses generation benchmarks).

\noindent
\textbf{Other Benchmarks of PM on LLMs}:
In \cite{DBLP:journals/corr/abs-2307-09909}, the authors assess the LLM-based translation of process mining questions proposed in \cite{DBLP:journals/jiis/BarbieriMSA23} to SQL statements.
They found that even advanced LLMs require the provision of database-specific and process-mining-specific domain knowledge, that requires prompt injection. Moreover, the questions that could
be translated to SQL statements are quantitative and do not assess the process-specific knowledge of the LLMs.
In \cite{DBLP:journals/corr/abs-2307-12701}, an initial comparison of two LLMs (GPT-4 and Google Bard) on different process mining tasks and implementation paradigms is performed.
The results lay the foundations for this paper, which proposes a much more comprehensive benchmark.
In \cite{DBLP:journals/corr/abs-2406-05506,DBLP:journals/corr/abs-2401-12846}, some benchmarks covering causal reasoning and explaining decision points in business processes are proposed.
In \cite{rebmann2024evaluating}, benchmarks for some semantics-aware process mining tasks, i.e. semantic anomaly detection and the prediction of the next activity, are provided along with strategies
to fine-tune the LLMs to improve their ability to execute the tasks.
Moreover, \cite{DBLP:journals/corr/abs-2406-13264} proposes a benchmark at the intersection between Robotic Process Automation and Business Process Management, evaluating the ability to exploit
workflows recorded in user screenshots for Business Process Management tasks.

\noindent
\textbf{LLMs-as-Judges}: As the outputs of LLMs are mainly textual, evaluating them in an automatic way is challenging. Some metrics have been proposed to evaluate how well an answer matches a ``ground truth''
provided by an human analyst\footnote{\url{https://mlflow.org/docs/latest/llms/llm-evaluate/index.html}}. However, their evaluation schema cannot be adapted to open-ended answers. Another option is to let
an LLM evaluate the answer (LLM-as-a-Judge) \cite{DBLP:conf/nips/ZhengC00WZL0LXZ23}. Using LLMs as judges, the evaluation can be tailored to the desired criteria and consider open-ended answers.
Some studies compared the scores given by humans and LLMs to a given set of questions/answers, finding a good alignment between humans and LLMs \cite{DBLP:conf/nips/ZhengC00WZL0LXZ23,thakur2024judging}.
However, some studies highlighted potential weaknesses due to bias \cite{DBLP:journals/corr/abs-2402-14016} and difficulty in following arbitrary evaluation directives \cite{dong2024can}.
In \cite{DBLP:journals/corr/abs-2309-16145}, it is shown that LLMs (as also humans) suffer from the Dunning-Kruger effect, underestimating/overestimating scores.
LLMs-as-Judges can be implemented with or without the provision of a ground truth \cite{DBLP:journals/corr/abs-2402-14860}. If no ground truth is provided, it is necessary that the judge LLM
would be able to i) respond correctly to the given inquiry; ii) identify errors and opportunities for improvement in the provided answer.
In ranking different LLMs without ground truth, there are other potential biases to consider.
In \cite{DBLP:journals/corr/abs-2305-17926}, it is highlighted how LLMs tend to prefer answers of similar LLMs.
The ``egocentric bias'' (LLMs preferring their own answers) is highlighted in \cite{DBLP:journals/corr/abs-2309-17012}.

\section{Benchmark}
\label{sec:benchmark}

In this section, we describe \emph{PM-LLM-Benchmark}, which is available at the address \url{https://github.com/fit-alessandro-berti/pm-llm-benchmark}.
First, in Section \ref{subsec:benchmarkCategoriesTasks}, we describe the categories of prompts included in the benchmarks.
Then, in Section \ref{subsec:benchmarkEvaluationStrategy}, we describe the evaluation strategy (LLM-as-a-Judge).

\subsection{Categories of Tasks}
\label{subsec:benchmarkCategoriesTasks}

Our benchmark measures how much an LLM is \emph{knowledgeable} and \emph{capable} in process mining. The capability is measured in the correct interpretation and production
of different process mining artifacts (traditional and object-centric event logs; procedural and declarative process models). We also evaluate the ability of the LLM to autonomously formulate hypotheses
over the event data or the process model. Moreover, as the goal of process mining is to assist data-driven decision-making, we aim to assess how much the LLM is able to identify biases starting from the event data.
We also want to assess the ability of Large Vision Language Models (LVLMs; so LLMs supporting visual prompts) to interpret popular visualizations and process mining diagrams.

\begin{table*}[!t]
\centering
\caption{Prompts included in the benchmark.}
\label{tab:promptsIncludedBenchmark}
\resizebox{\textwidth}{!}{
\begin{tabular}{|l|l|c|c|l|l|l|}
\hline
~ & \textbf{Prompt} & \textbf{Open} & \textbf{Requires DK} & \textbf{Task} & \textbf{Input Abstraction} & \textbf{Input Dataset} \\ \hline
\textbf{C1} & cat01\_01\_variants\_bpic2020\_rca & X & X & RCA & Variants & BPI2020 Domestic \\ \hline
\textbf{C1} & cat01\_02\_variants\_roadtraffic\_anomalies & X & X & Semantic AD & Variants & Road Traffic \\ \hline
\textbf{C1} & cat01\_03\_bpic2020\_var\_descr & X & X & Description & Variants & BPI2020 Domestic \\ \hline
\textbf{C1} & cat01\_04\_roadtraffic\_var\_descr & X & X & Description & Variants & Road Traffic \\ \hline
\textbf{C1} & cat01\_05\_bpic2020\_dfg\_descr & X & X & Description & DFG & BPI2020 Domestic \\ \hline
\textbf{C1} & cat01\_06\_roadtraffic\_dfg\_descr & X & X & Description & DFG & Road Traffic \\ \hline
\textbf{C1} & cat01\_07\_ocel\_container\_description & X & X & Description & OC-DFG & Logistics \\ \hline
\textbf{C1} & cat01\_08\_ocel\_order\_description & X & X & Description & OC-DFG & Order Management \\ \hline
\textbf{C1} & cat01\_09\_ocel\_container\_rca & X & X & RCA & OC-DFG & Logistics \\ \hline
\textbf{C1} & cat01\_10\_ocel\_order\_rca & X & X & RCA & OC-DFG & Order Management \\ \hline
\textbf{C2} & cat02\_01\_open\_event\_abstraction & X & X & Domain Knowledge & & \\ \hline
\textbf{C2} & cat02\_02\_open\_process\_cubes & X & X & Domain Knowledge & & \\ \hline
\textbf{C2} & cat02\_03\_open\_decomposition\_strategies & X & X & Domain Knowledge & & \\ \hline
\textbf{C2} & cat02\_04\_open\_trace\_clustering & X & X & Domain Knowledge & & \\ \hline
\textbf{C2} & cat02\_05\_open\_rpa & X & X & Domain Knowledge & & \\ \hline
\textbf{C2} & cat02\_06\_open\_anomaly\_detection & X & X & Domain Knowledge & & \\ \hline
\textbf{C2} & cat02\_07\_open\_process\_enhancement & X & X & Domain Knowledge & & \\ \hline
\textbf{C2} & cat02\_08\_closed\_process\_mining & & X & Domain Knowledge & & \\ \hline
\textbf{C2} & cat02\_09\_closed\_petri\_nets & & X & Domain Knowledge & & \\ \hline
\textbf{C3} & cat03\_01\_temp\_profile\_generation & X & X & Process Modeling & & \\ \hline
\textbf{C3} & cat03\_02\_declare\_generation & X & X & Process Modeling & & \\ \hline
\textbf{C3} & cat03\_03\_log\_skeleton\_generation & X & X & Process Modeling & & \\ \hline
\textbf{C3} & cat03\_04\_process\_tree\_generation & X & X & Process Modeling & & \\ \hline
\textbf{C3} & cat03\_05\_powl\_generation & X & X & Process Modeling & & \\ \hline
\textbf{C3} & cat03\_06\_temp\_profile\_discovery & X & X & Process Discovery & Variants & Road Traffic \\ \hline
\textbf{C3} & cat03\_07\_declare\_discovery & X & X & Process Discovery & Variants & Road Traffic \\ \hline
\textbf{C3} & cat03\_08\_log\_skeleton\_discovery & X & X & Process Discovery & Variants & Road Traffic \\ \hline
\textbf{C4} & cat04\_01\_bpmn\_xml\_tasks & X & X & Task List & BPMN XML & Running Example \\ \hline
\textbf{C4} & cat04\_02\_bpmn\_json\_description & X & X & Description & BPMN JSON & CCC19 \\ \hline
\textbf{C4} & cat04\_03\_bpmn\_simp\_xml\_description & X & X & Description & BPMN XML (simple) & CCC19 \\ \hline
\textbf{C4} & cat04\_04\_declare\_description & X & X & Description & DECLARE & BPI2020 Domestic \\ \hline
\textbf{C4} & cat04\_05\_declare\_anomalies & X & X & Semantic AD & DECLARE & BPI2020 Domestic \\ \hline
\textbf{C4} & cat04\_06\_log\_skeleton\_description & X & X & Description & Log Skeleton & BPI2020 Domestic \\ \hline
\textbf{C4} & cat04\_07\_log\_skeleton\_anomalies & X & X & Semantic AD & Log Skeleton & BPI2020 Domestic \\ \hline
\textbf{C5} & cat05\_01\_hypothesis\_bpic2020 & X & X & Hypothesis Generation & Variants & BPI2020 Domestic \\ \hline
\textbf{C5} & cat05\_02\_hypothesis\_roadtraffic & X & X & Hypothesis Generation & Variants & Road Traffic \\ \hline
\textbf{C5} & cat05\_03\_hypothesis\_bpmn\_json & X & X & Hypothesis Generation & BPMN JSON & CCC19 \\ \hline
\textbf{C5} & cat05\_04\_hypothesis\_bpmn\_simpl\_xml & X & X & Hypothesis Generation & BPMN XML (simple) & CCC19 \\ \hline
\textbf{C6} & cat06\_01\_renting\_attributes & & X & Discrimination Factors & Situation table & Renting (Fairness) \\ \hline
\textbf{C6} & cat06\_02\_hiring\_attributes & & X & Discrimination Factors & Situation table & Hiring (Fairness) \\ \hline
\textbf{C6} & cat06\_03\_lending\_attributes & & X & Discrimination Factors & Situation table & Lending (Fairness) \\ \hline
\textbf{C6} & cat06\_04\_hospital\_attributes & & X & Discrimination Factors & Situation table & Hospital (Fairness) \\ \hline
\textbf{C6} & cat06\_05\_renting\_prot\_comp & X & X & Comparison & Variants & Renting (Fairness) \\ \hline
\textbf{C6} & cat06\_06\_hiring\_prot\_comp & X & X & Comparison & Variants & Hiring (Fairness) \\ \hline
\textbf{C6} & cat06\_07\_lending\_prot\_comp & X & X & Comparison & Variants & Lending (Fairness) \\ \hline
\textbf{C6} & cat06\_08\_hospital\_prot\_comp & X & X & Comparison & Variants & Hospital (Fairness) \\ \hline
\textbf{C7} & cat07\_01\_dotted\_chart & X & X & Description & Visual & Road Traffic \\ \hline
\textbf{C7} & cat07\_02\_perf\_spectrum & X & X & Description & Visual & Road Traffic \\ \hline
\textbf{C7} & cat07\_03\_running-example & X & X & Description & Visual & Running Example \\ \hline
\textbf{C7} & cat07\_04\_credit-score & X & X & Description & Visual & Credit Score \\ \hline
\textbf{C7} & cat07\_05\_dfg\_ru & X & X & Description & Visual & Running Example \\ \hline
\textbf{C7} & cat07\_06\_process\_tree\_ru & X & X & Description & Visual & Running Example \\ \hline
\end{tabular}
}
\vspace{-5mm}
\end{table*}

The benchmark focuses on two implementation paradigms, i.e., the \emph{direct provision of insights} and \emph{code generation}.
Moreover, specific focus is given on process-mining-specific and process-specific \emph{domain knowledge}, which is required
for the considered prompts. Other available lists of process mining inquiries, such as the ones proposed in \cite{DBLP:journals/jiis/BarbieriMSA23},
are based on the generation of SQL statements but do not require process-specific domain knowledge.

Different categories of prompts (Table \ref{tab:promptsIncludedBenchmark}) are contained in the benchmark:
\begin{enumerate}[label=\textbf{C\arabic*}]
\item \textbf{General-purpose qualitative tasks}: The first category assesses the ability to describe processes, detect anomalies, and analyze root causes using DFG/variants abstractions of event logs. It also includes object-centric process mining artifacts for testing object-centric comprehension.
\item \textbf{Open/closed process mining domain knowledge questions}: The second category evaluates the process mining domain knowledge of the LLM, with open and closed questions about process mining and Petri nets.
\item \textbf{Process model generation}: The third category tests the ability to generate procedural (process trees, POWLs \cite{DBLP:conf/bpm/KouraniZ23}) and declarative process models (control-flow and temporal) for mainstream processes, and the ability to propose constraints given some process data.
\item \textbf{Process model understanding}: The fourth category assesses the understanding of proposed procedural (BPMN) and declarative (Log Skeleton and DECLARE \cite{DBLP:conf/bpm/Maggi13}) process models.
\item \textbf{Hypotheses generation}: The fifth category evaluates the ability to generate hypotheses over the proposed data and process models.
\item \textbf{Fairness assessment}: The sixth category tests the ability to identify event log attributes sensitive for fairness and compare protected and non-protected groups \cite{DBLP:conf/bpm/PohlBQA23}.
\item \textbf{Visual prompts}: The seventh category assesses the visual capabilities (if supported) of the LLM/LVLM.
\end{enumerate}

We tailored the benchmark to the level of comprehension and reasoning of currently available state-of-the-art LLMs (in particular, \emph{gpt-4o-20240513} and \emph{claude-3.5-sonnet}),
which can perform the tasks satisfactorily. Some mainstream tasks (for instance, applying the Alpha Miner algorithm, or checking the soundness of a Petri nets)
are still not supported effectively by state-of-the-art LLMs. Therefore, they have not been included in the current version of the benchmark,
which is comprehensive but not complete.
The prompts of the benchmark are ``static'' (i.e., stored in TXT files). The benchmark could be easily adapted in the future to contain more prompts and/or different categories of tasks.

The size of the prompt does not exceed $8K$ characters, ensuring their executability on any of the considered LLMs.
Among the considered LLMs, \emph{Llama3 70B Instruct} has the most restrictive context window (i.e., the number of tokens that can be provided).
Some state-of-the-art LLMs, such as \emph{Nemotron 340B} or \emph{Phi-3},
only support a baseline context window of $4K$, but 
we choose not to support that as it is too restrictive for process mining tasks.
For more difficult event logs or process models, a bigger context window ($32K$, $64K$, or $128K$) is preferable as it allows
to encode more information in the prompt.

\subsection{Evaluation Strategy}
\label{subsec:benchmarkEvaluationStrategy}

The challenge of evaluating textual outputs from LLMs in an objective manner is significant. Traditional metrics that compare answers to a human-provided ground truth are limited by their inability to be customized to specific evaluation criteria and to consider open-ended answers. This leads to the exploration of using LLMs as judges (LLMs-as-Judges), a method that offers the potential for more comprehensive evaluations.

As the output of LLMs for the proposed prompts is textual, we propose to use an advanced LLM (for instance, \emph{gpt-4o-20240513}) as judge \cite{DBLP:conf/nips/ZhengC00WZL0LXZ23}, assigning a score from 1.0 (minimum) to 10.0 (maximum) to each answer.

In this benchmark, LLMs-as-Judges are utilized without ground truth for two primary reasons:
\begin{itemize}
\item The open-ended nature of many inquiries, in particular \emph{process modeling} (C3) and \emph{automated hypotheses generation} (C5), which lack definitive answers.
\item The possibility that future LLM training sets could include the benchmark inquiries and their ground truths, potentially compromising the integrity of the benchmark by allowing ``cheating'' during the training phase of the LLM.
\end{itemize}

Potential limitations have been summarized in Section \ref{sec:relatedWork}.
In particular, LLMs-as-Judges are more reliable when there is a significant performance gap in favor of the judge LLM compared to the answering LLM.
Also, scoring similar-performing LLMs without ground truth is challenging and requires accounting for potential ``egocentric'' bias.
Moreover, high-performing LLMs tend to produce verbose and detailed outputs, which can result in a bias against more concise responses from lower-performing LLMs and vice-versa.

\begin{table*}[!t]
\centering
\caption{Scores between 1.0 and 10.0 ($mean \pm stddev$) of various LLMs in the proposed benchmark (using \emph{gpt-4o-20240513} as a judge).}
\label{tab:performanceScoresLlmsBenchmark}
\resizebox{0.95\textwidth}{!}{
\begin{tabular}{|c|c|c|c|}
\hline
\multicolumn{1}{|c|}{\textbf{Commercial LLMs}} & \multicolumn{1}{c|}{\textbf{Big Open-Source LLMs}} & \multicolumn{1}{c|}{\textbf{Small LLMs ($\leq$ 8GB)}} & \multicolumn{1}{c|}{\textbf{Tiny LLMs ($\leq$ 4GB)}} \\
\hline
claude-3.5-sonnet & Qwen v2.0 72B (instruct, fp16) & Qwen v2.0 7B (instruct, Q6K) & Mistral 7B v0.3 (instruct, Q3KS) \\
$8.4 \pm 0.7$ & $7.6 \pm 1.4$ & $6.5 \pm 2.2$ & $4.6 \pm 2.2$ \\
\hline
gpt-4o-20240513 (self) & WizardLM v2 8x22b (16b) & Mistral 7B v0.3 (instruct, Q6K) & Qwen v2.0 7B (instruct, Q2K) \\
$8.3 \pm 1.0$ & $7.5 \pm 1.7$ & $5.9 \pm 2.5$ & $4.6 \pm 2.1$ \\
\hline
gpt-4-turbo-20240409 & Mixtral v0.1 8x22b (instruct, 16b) & Llama 3 8B (instruct, Q6K) & Gemma v1.0 2B (instruct, Q6K) \\
$8.1 \pm 1.0$ & $7.5 \pm 1.6$ & $5.9 \pm 2.5$ & $4.0 \pm 2.6$ \\
\hline
claude-3-sonnet & Llama 3 70B (instruct, 16b) & WizardLM v2 7b (Q6K) & Qwen v2.0 1.5B (instruct, Q6K) \\
$7.8 \pm 1.4$ & $7.4 \pm 1.5$ & $5.9 \pm 2.3$ & $3.8 \pm 2.2$ \\
\hline
Google Gemini (20240528) & Mixtral v0.1 8x7b (instruct, 16b) & Gemma v2.0 9B (instruct, Q6K) & Qwen v2.0 0.5B (instruct, Q6K) \\
$7.5 \pm 1.7$ & $6.9 \pm 1.7$ & $5.7 \pm 2.9$ & $3.1 \pm 1.7$ \\
\hline
gpt-3.5-turbo-0125 & Llama 3 70B (instruct, Q4\_0) & CodeGemma v1.5 7B (instruct, Q6K) & Qwen 4B v1.5 (text, Q6K) \\
$7.1 \pm 1.8$ & $6.7 \pm 2.4$ & $4.9 \pm 2.6$ & $2.5 \pm 2.2$ \\
\hline
& Codestral 22B (Q6K) & Gemma v1.0 7B (instruct, Q6K) & \\
& $6.7 \pm 2.1$ & $4.5 \pm 2.6$ & \\
\cline{1-3}
& Llama 3 8B (instruct, 16b) & & \\
& $6.6 \pm 2.0$ & & \\
\cline{1-2}
& OpenChat 3.6 8B (16b) & & \\
& $6.5 \pm 1.7$ & & \\
\hline
\end{tabular}
}
\vspace{-5mm}
\end{table*}

\begin{table*}[!b]
\vspace{-7mm}
\centering
\caption{Scores between 1.0 and 10.0 ($mean \pm stddev$) for different model categories and question categories (using \emph{gpt-4o-20240513} as a judge).}
\label{tab:peformanceScoresModelQuestions}
\resizebox{0.95\textwidth}{!}{
\begin{tabular}{|c|c|c|c|c|c|c|c|}
\hline
\multicolumn{1}{|c|}{~} & \textbf{Commercial LLMs} & \textbf{Big Open-Source LLMs} & \textbf{Small LLMs ($\leq$ 8GB)} & \textbf{Tiny LLMs ($\leq$ 4GB)} \\
\hline
\textbf{C1} & $7.6 \pm 1.1$ & $7.0 \pm 1.5$ & $5.5 \pm 2.3$ & $3.2 \pm 1.6$ \\
\hline
\textbf{C2} & $8.7 \pm 0.6$ & $8.4 \pm 0.9$ & $8.5 \pm 0.9$ & $6.5 \pm 2.2$ \\
\hline
\textbf{C3} & $7.1 \pm 1.7$ & $5.7 \pm 2.0$ & $4.7 \pm 2.5$ & $2.4 \pm 1.5$ \\
\hline
\textbf{C4} & $7.7 \pm 1.5$ & $6.7 \pm 1.9$ & $3.7 \pm 2.0$ & $2.9 \pm 1.5$ \\
\hline
\textbf{C5} & $7.8 \pm 1.8$ & $7.3 \pm 1.6$ & $6.0 \pm 2.5$ & $4.6 \pm 2.2$ \\
\hline
\textbf{C6} & $8.1 \pm 1.2$ & $7.2 \pm 2.0$ & $5.0 \pm 2.2$ & $3.1 \pm 1.6$ \\
\hline
\textbf{C7} & $8.2 \pm 1.3$ & no supp. & no supp. & no supp. \\
\hline
\end{tabular}
}
\end{table*}

For the evaluation, the following procedure is followed for every prompt:
\begin{enumerate}
\item The prompt is provided to the LLM:
\begin{itemize}
\item Reported as-is for all the textual prompts.
\item For visual prompts (if supported by the given model), upload the image accompanied by the following textual prompt: \emph{Can you describe the provided visualization?}
\end{itemize}
\item The LLM's answer is persisted.
\item An expert LLM (LLM-as-a-Judge) is used to evaluate the output. Template:
\begin{itemize}
\item For textual prompts, \emph{Given the following question: \ldots How would you grade the following answer from 1.0 (minimum) to 10.0 (maximum)?}.
\item For visual prompts, upload the image to the LVLM and ask \emph{Given the attached image, how would you grade the following answer from 1.0 (minimum) to 10.0 (maximum)?}.
\end{itemize}
\end{enumerate}

\begin{table*}[!b]
\vspace{-5mm}
\caption{Scores between 1.0 and 10.0 ($mean \pm stddev$) for answering (rows) and evaluating (columns) LLM pairs.}
\label{tab:llmPairsScores}
\centering
\resizebox{\textwidth}{!}{
\begin{tabular}{|l|l|llll|}
\hline
\textbf{Answ./Eval.} & \textbf{gpt-4o-20240513} & \textbf{Llama3 70B Instr.} & \textbf{Mixtral 8x22B} & \textbf{Mixtral 8x7B} & \textbf{Qwen2 72B Instr.} \\ \hline
Llama3 70B Instr. & $7.4 \pm 1.5$ & $\mathbf{8.8 \pm 0.5}$ & $8.6 \pm 1.4$ & $8.7 \pm 1.3$ & $7.7 \pm 2.4$ \\ \hline
WizardLM-2-8x22B & $7.5 \pm 1.7$ & $7.2 \pm 3.0$ & $7.2 \pm 2.7$ & $8.3 \pm 1.3$ & $7.1 \pm 2.9$ \\ \hline
Mixtral 8x22B & $7.5 \pm 1.6$ & $7.8 \pm 2.3$ & $7.5 \pm 3.1$ & $8.5 \pm 1.3$ & $7.6 \pm 2.4$ \\ \hline
Mixtral 8x7B & $6.9 \pm 1.7$ & $7.7 \pm 2.3$ & $7.4 \pm 2.9$ & $8.1 \pm 1.7$ & $7.1 \pm 2.8$ \\ \hline
Qwen2 72B Instr. & $7.6 \pm 1.4$ & $7.9 \pm 2.3$ & $8.6 \pm 1.3$ & $8.3 \pm 1.9$ & $\mathbf{8.0 \pm 2.5}$ \\ \hline
\end{tabular}
}
\end{table*}

\begin{table*}[!t]
\centering
\caption{Scores between 1.0 and 10.0 ($mean \pm stddev$) for different model categories and single prompts (using \emph{gpt-4o-20240513} as a judge).}
\label{tab:performanceSinglePrompts}
\resizebox{0.65\textwidth}{!}{
\begin{tabular}{|c|c|c|c|c|}
\hline
\multicolumn{1}{|c|}{\textbf{Prompt}} & \textbf{Comm.} & \textbf{Big OS} & \textbf{Small} & \textbf{Tiny} \\
\hline
cat01\_01\_variants\_bpic2020\_rca & $8.6 \pm 0.4$ & $6.9 \pm 1.4$ & $5.7 \pm 2.9$ & $3.2 \pm 1.7$ \\
\hline
cat01\_02\_variants\_roadtraffic\_anomalies & $6.5 \pm 1.6$ & $6.1 \pm 1.8$ & $4.3 \pm 2.4$ & $2.2 \pm 0.9$ \\
\hline
cat01\_03\_bpic2020\_var\_descr & $7.8 \pm 1.6$ & $7.5 \pm 0.9$ & $7.0 \pm 1.3$ & $3.4 \pm 1.4$ \\
\hline
cat01\_04\_roadtraffic\_var\_descr & $8.3 \pm 0.6$ & $7.4 \pm 0.9$ & $6.8 \pm 1.4$ & $3.2 \pm 1.6$ \\
\hline
cat01\_05\_bpic2020\_dfg\_descr & $8.2 \pm 0.6$ & $8.2 \pm 0.7$ & $7.4 \pm 1.8$ & $3.8 \pm 2.1$ \\
\hline
cat01\_06\_roadtraffic\_dfg\_descr & $7.8 \pm 0.6$ & $6.4 \pm 1.8$ & $5.1 \pm 2.1$ & $3.9 \pm 1.9$ \\
\hline
cat01\_07\_ocel\_container\_description & $7.2 \pm 0.9$ & $6.6 \pm 1.5$ & $4.9 \pm 1.2$ & $4.0 \pm 1.3$ \\
\hline
cat01\_08\_ocel\_order\_description & $7.7 \pm 0.7$ & $6.8 \pm 1.4$ & $4.8 \pm 2.3$ & $3.5 \pm 1.6$ \\
\hline
cat01\_09\_ocel\_container\_rca & $7.2 \pm 1.2$ & $7.5 \pm 0.8$ & $5.4 \pm 2.1$ & $2.6 \pm 0.9$ \\
\hline
cat01\_10\_ocel\_order\_rca & $7.2 \pm 0.7$ & $6.7 \pm 1.8$ & $3.5 \pm 1.3$ & $2.5 \pm 1.0$ \\
\hline
cat02\_01\_open\_event\_abstraction & $8.7 \pm 0.4$ & $8.4 \pm 0.6$ & $8.7 \pm 0.4$ & $6.6 \pm 1.8$ \\
\hline
cat02\_02\_open\_process\_cubes & $8.9 \pm 0.2$ & $8.4 \pm 0.5$ & $8.5 \pm 0.5$ & $5.8 \pm 1.8$ \\
\hline
cat02\_03\_open\_decomposition\_strategies & $8.7 \pm 0.9$ & $8.5 \pm 0.3$ & $8.6 \pm 0.3$ & $7.2 \pm 1.2$ \\
\hline
cat02\_04\_open\_trace\_clustering & $8.5 \pm 0.7$ & $8.4 \pm 0.5$ & $8.7 \pm 0.7$ & $7.7 \pm 0.9$ \\
\hline
cat02\_05\_open\_rpa & $9.1 \pm 0.2$ & $8.8 \pm 0.4$ & $9.0 \pm 0.5$ & $8.2 \pm 1.2$ \\
\hline
cat02\_06\_open\_anomaly\_detection & $8.6 \pm 0.5$ & $8.7 \pm 0.4$ & $8.1 \pm 1.0$ & $8.0 \pm 1.4$ \\
\hline
cat02\_07\_open\_process\_enhancement & $8.9 \pm 0.3$ & $8.7 \pm 0.4$ & $8.7 \pm 0.7$ & $6.4 \pm 1.0$ \\
\hline
cat02\_08\_closed\_process\_mining & $8.8 \pm 0.6$ & $8.0 \pm 1.1$ & $8.8 \pm 0.7$ & $6.0 \pm 2.7$ \\
\hline
cat02\_09\_closed\_petri\_nets & $8.1 \pm 0.6$ & $7.2 \pm 1.8$ & $6.9 \pm 1.3$ & $2.9 \pm 1.2$ \\
\hline
cat03\_01\_temp\_profile\_generation & $9.0 \pm 0.1$ & $7.8 \pm 0.9$ & $7.8 \pm 1.2$ & $3.4 \pm 1.6$ \\
\hline
cat03\_02\_declare\_generation & $6.8 \pm 1.5$ & $5.8 \pm 1.2$ & $4.8 \pm 1.4$ & $2.7 \pm 0.9$ \\
\hline
cat03\_03\_log\_skeleton\_generation & $8.2 \pm 1.2$ & $7.3 \pm 1.2$ & $5.8 \pm 1.7$ & $3.2 \pm 1.6$ \\
\hline
cat03\_04\_process\_tree\_generation & $7.4 \pm 1.2$ & $5.6 \pm 2.0$ & $5.6 \pm 1.6$ & $3.4 \pm 1.4$ \\
\hline
cat03\_05\_powl\_generation & $6.6 \pm 1.2$ & $6.5 \pm 1.5$ & $6.9 \pm 2.0$ & $2.5 \pm 1.4$ \\
\hline
cat03\_06\_temp\_profile\_discovery & $5.5 \pm 1.8$ & $4.2 \pm 1.7$ & $2.4 \pm 0.7$ & $1.7 \pm 0.7$ \\
\hline
cat03\_07\_declare\_discovery & $7.4 \pm 1.3$ & $4.3 \pm 1.4$ & $1.9 \pm 1.0$ & $1.0 \pm 0.0$ \\
\hline
cat03\_08\_log\_skeleton\_discovery & $5.8 \pm 1.2$ & $4.1 \pm 1.4$ & $2.2 \pm 1.1$ & $1.2 \pm 0.6$ \\
\hline
cat04\_01\_bpmn\_xml\_tasks & $9.2 \pm 0.6$ & $7.8 \pm 3.1$ & $1.3 \pm 0.5$ & $1.7 \pm 0.7$ \\
\hline
cat04\_02\_bpmn\_json\_description & $7.9 \pm 1.0$ & $6.3 \pm 2.0$ & $2.9 \pm 1.5$ & $2.8 \pm 2.1$ \\
\hline
cat04\_03\_bpmn\_simp\_xml\_description & $8.2 \pm 0.6$ & $6.7 \pm 1.2$ & $3.6 \pm 1.9$ & $2.6 \pm 1.4$ \\
\hline
cat04\_04\_declare\_description & $7.3 \pm 1.1$ & $7.1 \pm 1.1$ & $5.7 \pm 1.6$ & $3.6 \pm 1.4$ \\
\hline
cat04\_05\_declare\_anomalies & $6.8 \pm 2.0$ & $5.9 \pm 1.6$ & $3.4 \pm 0.9$ & $3.2 \pm 1.3$ \\
\hline
cat04\_06\_log\_skeleton\_description & $7.9 \pm 1.3$ & $6.6 \pm 1.4$ & $5.2 \pm 1.9$ & $3.3 \pm 1.7$ \\
\hline
cat04\_07\_log\_skeleton\_anomalies & $6.5 \pm 1.3$ & $6.5 \pm 1.6$ & $4.0 \pm 1.9$ & $2.9 \pm 0.7$ \\
\hline
cat05\_01\_hypothesis\_bpic2020 & $8.2 \pm 0.4$ & $7.4 \pm 1.4$ & $7.6 \pm 0.7$ & $4.6 \pm 1.7$ \\
\hline
cat05\_02\_hypothesis\_roadtraffic & $8.1 \pm 1.0$ & $6.6 \pm 1.7$ & $6.8 \pm 2.2$ & $4.7 \pm 2.3$ \\
\hline
cat05\_03\_hypothesis\_bpmn\_json & $7.5 \pm 1.6$ & $7.7 \pm 1.2$ & $5.2 \pm 2.2$ & $4.5 \pm 2.3$ \\
\hline
cat05\_04\_hypothesis\_bpmn\_simpl\_xml & $7.2 \pm 2.9$ & $7.4 \pm 1.9$ & $4.6 \pm 2.9$ & $4.5 \pm 2.4$ \\
\hline
cat06\_01\_renting\_attributes & $8.8 \pm 0.6$ & $8.2 \pm 1.4$ & $4.1 \pm 2.0$ & $2.5 \pm 1.1$ \\
\hline
cat06\_02\_hiring\_attributes & $8.9 \pm 0.7$ & $8.6 \pm 0.8$ & $6.1 \pm 2.4$ & $5.0 \pm 2.2$ \\
\hline
cat06\_03\_lending\_attributes & $8.3 \pm 0.5$ & $8.3 \pm 1.1$ & $7.1 \pm 2.3$ & $2.8 \pm 1.2$ \\
\hline
cat06\_04\_hospital\_attributes & $8.8 \pm 0.5$ & $8.4 \pm 0.7$ & $6.3 \pm 1.8$ & $4.4 \pm 2.1$ \\
\hline
cat06\_05\_renting\_prot\_comp & $7.5 \pm 1.2$ & $5.8 \pm 1.8$ & $3.9 \pm 1.1$ & $2.0 \pm 0.0$ \\
\hline
cat06\_06\_hiring\_prot\_comp & $7.2 \pm 1.4$ & $6.1 \pm 2.5$ & $4.4 \pm 1.7$ & $2.9 \pm 0.9$ \\
\hline
cat06\_07\_lending\_prot\_comp & $8.3 \pm 0.6$ & $6.5 \pm 1.8$ & $3.6 \pm 1.6$ & $2.3 \pm 0.5$ \\
\hline
cat06\_08\_hospital\_prot\_comp & $7.0 \pm 1.6$ & $5.8 \pm 1.8$ & $4.4 \pm 1.6$ & $2.7 \pm 0.9$ \\
\hline
cat07\_01\_dotted\_chart & $8.3 \pm 0.7$ & - & - & - \\
\hline
cat07\_02\_perf\_spectrum & $8.5 \pm 0.8$ & - & - & - \\
\hline
cat07\_03\_running-example & $8.9 \pm 0.5$ & - & - & - \\
\hline
cat07\_04\_credit-score & $8.6 \pm 0.6$ & - & - & - \\
\hline
cat07\_05\_dfg\_ru & $7.4 \pm 2.3$ & - & - & - \\
\hline
cat07\_06\_process\_tree\_ru & $7.8 \pm 1.0$ & - & - & - \\
\hline
\end{tabular}
}
\vspace{-5mm}
\end{table*}

\subsection{Benchmarking Scripts}
\label{subsec:benchmarkingScripts}

The provided benchmark can be executed manually (first, an LLM is used to answer the questions, and its answers are evaluated by another LLM).
We also provide some scripts to automate the execution of the benchmark.
In particular, the Python script \textbf{answer.py} can be used to execute the prompts against any LLM supporting the OpenAI APIs,
while \textbf{evaluation.py} can be used to evaluate the answers using an LLM as the judge. The configuration parameters could be set up inside
the two scripts.

\section{Benchmark Results}
\label{sec:benchmarkResults}

We executed the proposed benchmark against current state-of-the-art LLMs.
In Table \ref{tab:performanceScoresLlmsBenchmark}, we collect the evaluation results (using \emph{gpt-4o-20240513} as a judge) for different LLMs of different sizes.
In particular, we divide between \emph{commercial models} (usually ranging hundreds of billions of parameters),
\emph{big open-source LLMs}, \emph{small LLMs} ($\leq$ 8 GB of RAM), and \emph{tiny LLMs} ($\leq$ 4 GB of RAM).

We tested some LLMs with different levels of quantization.
Quantization refers to the process of reducing the precision of the model's weights and activations from higher bit-widths (such as 16-bit floating point) to lower bit-widths (such as 8-bit or even lower). This technique aims to decrease the model's memory footprint and computational requirements, making it more efficient to run on hardware with limited resources. While quantization can lead to a slight loss in model accuracy, it often significantly improves the model's speed and reduces power consumption, enabling more practical and scalable deployment of LLMs.

In general, we see that commercial and big open-source models can perform process mining tasks adequately well.
The winners among commercial models seem to be \emph{gpt-4o-20240513} and \emph{claude-3.5-sonnet}.
The best of the open-source models seem to be \emph{Qwen2 72B Instruct} (occupying $145 GB$ of memory at full quantization).
Some small LLMs also report an overall sufficient score. For instance, \emph{Qwen2 7B Instruct} at \emph{Q6K} quantization
is process-mining-capable while occupying just $6.3 GB$ of memory. However, tiny LLMs are still inadequate for process mining tasks.

Overall, we found that aggressive quantization has a severe impact on LLMs' abilities in the proposed benchmark. While we do not have precise explanations for this phenomenon,
the benchmarks' prompt requires significant attention to some elements of the input prompt (e.g., the semantic anomalies), and quantization impacts such ability.

In Table \ref{tab:peformanceScoresModelQuestions}, we report the scores for every model category and question category.
In Table \ref{tab:performanceSinglePrompts}, the scores for every prompt of the benchmark are averaged over a model category.

We notice that question category \textbf{C2} (open/closed process mining domain knowledge questions) is answered adequately by all the model categories.
Also, currently, only commercial models could answer to \textbf{C7} (visual prompts).
Surprisingly, most of the considered LLMs could automatically generate hypotheses over the provided process models and event data (\textbf{C5}).
Categories \textbf{C3} (process model generation) and \textbf{C4} (process model understanding) have high variance between the different model categories.
While commercial and big open-source LLMs can perform the tasks adequately, small and tiny models fail to generate/understand the information provided in the prompt.
Categories \textbf{C1} (general-purpose qualitative tasks) and \textbf{C6} (fairness assessment) have similar outcomes. Commercial and big open-source LLMs successfully execute
such tasks, while small/tiny LLMs usually fail.

To further assess the validity of LLMs-as-Judges, we report in Table \ref{tab:llmPairsScores} a cross-validation based on
five open-source answering models and five evaluation models (including \emph{gpt-4o-20240513}). We see that only \emph{Llama 3 70B Instruct}
display the egocentric bias. The models tend to agree on the scores. The smallest considered model (\emph{Mixtral 8x7B}) reports lower scores
than its bigger counterpart (\emph{Mixtral 8x22B}) and other state-of-the-art models (\emph{Llama 3 70B Instruct} and \emph{Qwen2 72B Instruct}).
Notably, \emph{WizardLM2 8x22B} achieves lower overall scores than \emph{Mixtral 8x7B}. This could be explained by the model
being a fine-tuned version of \emph{Mixtral 8x22B} favoring more verbose responses.
Overall, only \emph{Qwen2 72B Instruct} and \emph{gpt-4o-20240513} favor Qwen2 over Llama3, while Qwen2 is an overall better-performing
model in general-purpose benchmarks. As the models and responses become more advanced, it becomes complicated for lower-performing models to judge
the differences between them. Using the less advanced \emph{Mixtral 8x7B}, the ``plateauing'' of the scores becomes evident,
highlighting the importance of using advanced LLMs as judges.
Therefore, Table \ref{tab:llmPairsScores} justifies LLMs-as-Judges, but points to the importance of choosing advanced LLMs for the task.

\section{Future Benchmarking Strategies}
\label{subsec:futureBenchmarkingStrategies}

\noindent
\textbf{Benchmarking Retrieval-Augmented Generation (RAG)}: while LLMs have been trained on generic process-specific knowledge, the implementation of business processes in real-life organizations
could be different. Therefore, Retrieval-Augmented Generation (RAG) techniques have been used to dynamically inject process-specific knowledge to the prompts provided by process mining analysts \cite{DBLP:journals/corr/abs-2310-14735}.
As the correct retrieval of the information is crucial for the quality of the answers, benchmarks should also assess the capabilities of the RAG pipeline.

\noindent
\textbf{Benchmarking LLM-Based Agent Crews}:
Our benchmark focuses on executing a prompt, recording the answer, and evaluating it. The \emph{agents crew} paradigm \cite{DBLP:journals/corr/abs-2402-01680} involves workflows with specialized or role-specific LLMs performing analytical tasks. For instance, estimating the level of discrimination in an event log involves identifying the protected group and comparing differences between groups. An agents crew could consist of one agent creating the SQL statement to filter the protected group and another specializing in process comparison. Each agent's performance is crucial for accurate benchmarking, as evaluating only the final output may be misleading.

\noindent
\textbf{Benchmarking Hypotheses Refinement}:
Hypotheses generation involves creating and verifying hypotheses against event data. If a hypothesis is invalid, the LLM refines it based on feedback. This process may lead to specific hypotheses that perform poorly on imseem data. Thus, it is important to test the entire hypothesis feedback and verification cycle, not just hypotheses generation.

\noindent
\textbf{Data Generation with Relevant Semantic Anomalies or Root Causes}:
Our benchmark uses publicly available event data and process models. Including these datasets in LLM training data, even without ground truth answers, can lead to higher benchmark scores. Therefore, it is important to dynamically generate novel event data for LLM benchmarking. Current simulation solutions lack the semantic understanding needed to generate data suitable for process mining assessment. LLMs should be considered for the goal of generating data with various semantic anomalies and root causes.

\section{Conclusion}
\label{sec:conclusion}

We proposed \emph{PM-LLM-Benchmark}, a benchmark for process mining on LLMs, utilizing LLMs-as-Judges for scalable evaluation. This benchmark does not provide ground truth answers, relying instead on the "judge" LLM's capabilities, addressing the open-ended nature of the inquiries.
The benchmark includes various task categories to assess LLMs' abilities in process mining, modeling, and comprehension. Applied to several commercial and open-source LLMs, we found most models perform well in process mining tasks, with bigger models achieving higher scores. Smaller models ($\leq$ 8 GB or $\leq$ 4 GB of RAM) struggle with complex tasks.
We also suggest improvements in benchmarking strategies, including enhanced hypotheses generation, agent-crew-based benchmarks, and dynamic dataset generation with semantic anomalies.
To our knowledge, this is the first general-purpose process mining benchmark focusing on different implementation strategies. While it has limitations in evaluating well-performing LLMs, it is a useful tool for assessing smaller LLMs and evaluating the current state of PM-on-LLMs.

\bibliographystyle{splncs04}
\bibliography{references}

\end{document}